\documentclass[lettersize,journal]{IEEEtran}
\usepackage{graphicx}
\usepackage{amsmath, amssymb}

\usepackage{caption, subcaption}
\usepackage{multirow}
\usepackage{hyperref}
\usepackage{cite}
\begin{document}
\title{WaveMamba: Spatial-Spectral Wavelet Mamba for Hyperspectral Image Classification}
\author{Muhammad Ahmad, Muhammad Usama, Manuel Mazzara, Salvatore Distefano
\thanks{This work is partially funded by European Union - Next generation EU - PNRR - Missione 4, Componente 2, Investimento 1.1 - Bando PRIN 2022 PNRR - Decreto Direttoriale n. 1409 del 14-09-2022 - Progetto RESILIENT, CUP J53D23015040001, project id. P2022S4TTP, and Next generation EU - PNRR, SERICS – ``SECURITY AND RIGHTS IN THE CYBERSPACE'' project 3D-SEECSDE, CUP J33C22002810001, project id. PE00000014.}
\thanks{M. Ahmad and S. Distefano are with the Dipartimento di Matematica e Informatica---MIFT, University of Messina, Messina 98121, Italy. (e-mail: mahmad00@gmail.com).}
\thanks{M. Usama is with the Department of Computer Science, National University of Computer and Emerging Sciences, Islamabad, Chiniot-Faisalabad Campus, Chiniot 35400, Pakistan.}
\thanks{M. Mazzara is with the Institute of Software Development and Engineering, Innopolis University, Innopolis, 420500, Russia.}
\thanks{The source code is available upon request from the corresponding author.}
}
\markboth{IEEE Geoscience and Remote Sensing Letters}
{}
\maketitle
\begin{abstract}
Hyperspectral Imaging (HSI) has proven to be a powerful tool for capturing detailed spectral and spatial information across diverse applications. Despite the advancements in Deep Learning (DL) and Transformer architectures for HSI classification, challenges such as computational efficiency and the need for extensive labeled data persist. This paper introduces WaveMamba, a novel approach that integrates wavelet transformation with the spatial-spectral Mamba architecture to enhance HSI classification. WaveMamba captures both local texture patterns and global contextual relationships in an end-to-end trainable model. The Wavelet-based enhanced features are then processed through the state-space architecture to model spatial-spectral relationships and temporal dependencies. The experimental results indicate that WaveMamba surpasses existing models, achieving an accuracy improvement of 4.5\% on the University of Houston dataset and a 2.0\% increase on the Pavia University dataset.
\end{abstract}
\section{Introduction}

\IEEEPARstart{H}{yperspectral} Imaging (HSI) is a robust technique for capturing extensive spectral information across a continuous range of wavelengths. This capability facilitates detailed analysis and monitoring in various applications \cite{ahmad2021hyperspectral}. While HSI excels in providing detailed spatial and spectral data, high spectral resolution sensors may face challenges in achieving optimal spatial resolution in complex environments. As a result, the extensive spectral data presents both challenges and opportunities for effective classification \cite{10647404}.

Recent advances in Deep Learning (DL) have significantly improved HSI classification performance \cite{10332250, 10440363, 101016, DING2023119858}. Convolutional Neural Networks (CNNs) are particularly effective at extracting essential spatial and spectral features, enabling accurate classification \cite{hong2023decoupled, ahmad2020fast, 10423094}. However, their reliance on local receptive fields and substantial labeled data highlights the need for more advanced architectures to exploit global context \cite{hong2024spectralgpt, 10685113, 10433668}. The introduction of Transformer architecture has further enhanced HSI classification by effectively modeling long-range dependencies and global context, often outperforming traditional DL models \cite{hong2022spectralformer, 10681622, 10463068}. Despite these improvements, Transformers face challenges related to computational efficiency and the requirement for extensive labeled data, which limits their practical application \cite{10604879}.

Recently, the Mamba architecture, inspired by control theory's State Space Model, has achieved linear complexity scaling with sequence length, thereby enhancing HSI classification efficiency \cite{gu2023mamba}. Yao et al. \cite{yao2024spectralmamba} introduced SpectralMamba, which integrates Mamba with DL techniques through a gated spatial-spectral merging (GSSM) module and a piece-wise sequential scanning (PSS) strategy. Huang et al. \cite{huang2024spectral} proposed the spatial-spectral Mamba (SSMamba) architecture, utilizing spectral-spatial token generation and multiple stacked Mamba blocks for feature fusion, although further optimization is required. Wang et al. \cite{wang2024s} developed the $S^2$ Mamba architecture, which merges spatial and spectral features using a spatial-spectral mixture gate to enhance accuracy, but further optimization is necessary for complex interactions. He et al. \cite{he20243dss} introduced the 3D spectral-spatial Mamba (3DSS-Mamba) architecture, capturing global dependencies while maintaining linear complexity; however, optimization is still needed for high-dimensional data.

To address these challenges, this paper proposes WaveMamba, a novel approach that combines wavelet transformation with the spatial-spectral Mamba architecture for improved HSI classification. WaveMamba effectively captures both local texture patterns and global contextual relationships in an end-to-end trainable model. The wavelet-enhanced features are processed through the state-space architecture, modeling spatial-spectral relationships and temporal dependencies. This results in more accurate classification compared to traditional DL methods. By extracting wavelet-based spatial-spectral features from HSI data and integrating them into the Mamba architecture, WaveMamba captures local and global relationships, leading to improved classification accuracy. This paper makes the following contributions:

\begin{figure*}[!hbt]
    \centering
    \includegraphics[width=0.98\textwidth]{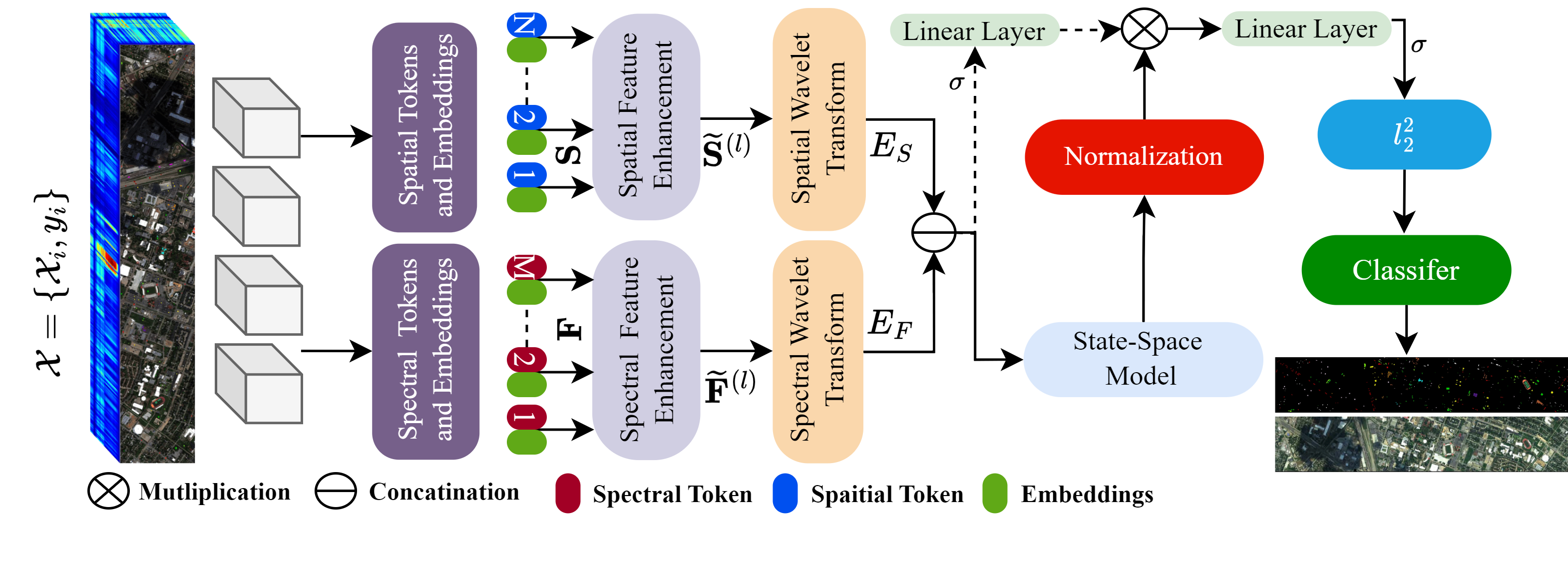}
    \caption{The HSI cube is divided into overlapping 3D patches further split into spectral and spatial tokens, enhanced using a gate mechanism, and transformed with Haar wavelets into four subbands that capture different frequency components and spatial features. The subbands are concatenated into a new 3D representation, which the Mamba architecture uses to capture spatial-spectral relationships and temporal dependencies.}
    \label{Fig1}
\end{figure*}

This work introduces WaveMamba, a method that integrates wavelet transformation with spatial-spectral Mamba for HSI classification. By leveraging wavelet-based multi-resolution analysis, WaveMamba enhances the interaction between spectral and spatial information, capturing a comprehensive range of data critical for accurate classification. The use of overlapping 3D patches along with wavelets improves local feature extraction, capturing fine-grained details. Additionally, a gate mechanism with fully connected dense and non-linear activation layers refines spatial and spectral tokens, enhancing feature representation.

This work demonstrates that combining wavelet-based feature extraction with the Mamba architecture's spatial-spectral relationship modeling achieves superior accuracy compared to traditional DL, Transformer, and Mamba methods. Furthermore, incorporating the State-Space Model captures temporal dependencies within HSI data, while $L_2$ regularization in the final classification stage ensures model simplicity and generalizability, reducing overfitting and enhancing performance on unseen data.

\section{Proposed Methodology}
\label{Meth}

An HSI cube $\mathcal{X} = \{\mathcal{X}_i, y_i\} \in \mathcal{R}^{(H \times W \times C)}$ consists of spectral vectors $\mathcal{X}_i = \{\mathcal{X}_{i,1}, \mathcal{X}_{i,2}, \mathcal{X}_{i,3}, \dots, \mathcal{X}_{i,C}\}$, and $y_i$ corresponding class label of $\mathcal{X}_i$. The cube is divided into overlapping 3D patches centered at coordinates $(\alpha, \beta)$, each covering $P \times P$ pixels across $C^*$ bands. The total number of patches $N$ is $(H-P+1) \times (W-P+1)$, where a patch $P_{\alpha, \beta}$ spans $\alpha \pm \frac{P-1}{2}$ and $\beta \pm \frac{P-1}{2}$ in spatial dimensions. The input shape is $(N, P, P, C^*)$, with $C^*$ representing the reduced number of bands. These patches are divided into spectral and spatial components to generate spectral tokens $\mathbf{F}$ and spatial tokens $\mathbf{S}$. By transforming 2D spatial data into $\mathbf{S}$ and 1D spectral data into $\mathbf{F}$, the model captures complex spatial-spectral relationships. The spatial and spectral tokens, $\mathbf{S}$ and $\mathbf{F}$, are generated as follows:

\begin{equation}
    \mathbf{S} = [\mathbf{s}_1, \mathbf{s}_2, \ldots, \mathbf{s}_C] \in \mathbb{R}^{B \times (HW) \times C}
\end{equation}
\begin{equation}
    \mathbf{F} = [\mathbf{f}_1, \mathbf{f}_2, \ldots, \mathbf{f}_{HW}] \in \mathbb{R}^{B \times (HW) \times C}
\end{equation}

Spatial-spectral feature enhancement is subsequently performed which refines the HSI data features by processing $\mathbf{S}$ and $\mathbf{F}$ tokens using a "spatial and spectral gate" mechanism. This mechanism involves fully connected dense layers, non-linear activation functions, and reshaped layers. The spatial gate enhances $\mathbf{S}$, and the spectral gate enhances $\mathbf{F}$. This process results in improved spatial and spectral token representations for subsequent analysis.

\begin{equation}
    \widetilde{\mathbf{S}}^{(l)} = \mathbf{S}^{(l)} \odot \sigma(\mathbf{W}_s \mathbf{c} + \mathbf{b}_s)
\end{equation}
\begin{equation}
    \widetilde{\mathbf{F}}^{(l)} = \mathbf{F}^{(l)} \odot \sigma(\mathbf{W}_f \mathbf{c} + \mathbf{b}_f)
\end{equation}
where $W_s$ and $W_f$ are learned weights and $\sigma$ is the sigmoid function. The enhanced spatial-spectral feature map $\widetilde{\mathbf{S}}^{(l)}$ and $\widetilde{\mathbf{F}}^{(l)}$ are transformed into $\mathbf{\hat{S}}^{(l)}$ and $\mathbf{\hat{F}}^{(l)}$ i.e., $\mathbf{\hat{S}}^{(l)}$ $\text{and}$ $\mathbf{\hat{F}}^{(l)} \in \mathcal{R}^{H \times W \times C^*}$ decomposed into 4 wavelet\footnote{We leverage wavelets in our methodology to capture multi-scale features, improving the representation of both spatial and spectral information} sub-channels which are down-sampled through wavelet transformation. Note that, here we used the classical Haar wavelet as expressed in \cite{liu2018multi, fujieda2018wavelet, 10399798}. Concretely, the wavelet transformation is applied using a lowpass filter, denoted as $f_l = \left(\frac{1}{\sqrt 2},\frac{1}{\sqrt 2}\right)$, and a highpass filter, denoted as $f_h = \left(\frac{1}{\sqrt 2},-\frac{1}{\sqrt 2}\right)$, along the rows of the input spatial and spectral features $\mathbf{\hat{S}}^{(l)}$ and $\mathbf{\hat{F}}^{(l)}$. This process results in the creation of four sub-channels, namely $\mathbf{\hat{S}}^{(l)}_l$, $\mathbf{\hat{S}}^{(l)}_h$ (Spatial) and $\mathbf{\hat{F}}^{(l)}_l$, $\mathbf{\hat{F}}^{(l)}_h$ (Spectral). Subsequently, the same lowpass filter $f_l$ and highpass filter $f_h$ are employed, this time along the columns of the derived sub-channels. This leads to the formation of eight sub-channels in total: $\mathbf{\hat{S}}^{(l)}_{ll}$, $\mathbf{\hat{S}}^{(l)}_{lh}$, $\mathbf{\hat{S}}^{(l)}_{hl}$, and $\mathbf{\hat{S}}^{(l)}_{hh}$ for spatial information and similarly $\mathbf{\hat{F}}^{(l)}_{ll}$, $\mathbf{\hat{F}}^{(l)}_{lh}$, $\mathbf{\hat{F}}^{(l)}_{hl}$, and $\mathbf{\hat{F}}^{(l)}_{hh}$ for spectral information. Each of these wavelet subbands can be viewed as a down-sampled version of the original input $\widetilde{\mathbf{S}}^{(l)}$ and $\widetilde{\mathbf{F}}^{(l)}$. They collectively preserve all the input details without any loss of information. 

The wavelet features are linearly transformed into the state-space model\footnote{State-space models represent systems through a set of hidden states that evolve over time. In our model, state-space modeling helps capture spatial-spectral dependencies.}. Given a sequence of wavelet features $O = (E_1, E_2, E_3, \dots, E_T)$, the state transition is computed as:
\begin{equation}
h_t = \text{ReLU}(W_{\text{transition}} h_{t-1} + W_{\text{update}} E_t)
\end{equation}
where $W_{\text{transition}}$ and $W_{\text{update}}$ are learned weights, and $h_t$ is the hidden state at time $t$. The final output is obtained by applying a linear classifier to $h_t$:
\begin{equation}
y = \sigma(h_t W_{\text{classifier}})
\end{equation}
where $|W_{\text{classifier}}|_2^2$ is the squared $l_2$ norm of $W_{\text{classifier}}$. The regularization coefficient $\lambda = 0.01$ controls the strength of regularization, with $\sigma$ denoting the sigmoid function for classification. $l_2$ regularization penalizes large values in $W_{\text{classifier}}$, promoting simpler models and preventing overfitting. This approach allows WaveMamba to effectively utilize both spatial and spectral information, improving its classification of HSI data by integrating insights from both modalities.

\section{Experimental Results and Discussion}
\label{ERD}

For all the following results, WaveMamba trained over 50 epochs with Adam optimizer and a 0.001 learning rate, effectively learned patterns using a mini-batch size of 256, Mamba block embeddings of 64, state space dimensions of 128, and a 0.1 dropout rate with $l_2^2$ regularization. The model weights were randomly initialized and fine-tuned, achieving improved performance.

\subsection{Patch Size and Train-Validation-Test Samples}

The effects of training sample percentage and patch size on WaveMamba performance were evaluated to determine the optimal configuration. Table \ref{Tab2} shows that varying training percentages with a fixed $4 \times 4$ patch size and testing different patch sizes reveal that smaller patches capture fine details but may overfit due to noise sensitivity. Larger patches, in contrast, capture global features, improving robustness and generalization as sample sizes increase. Optimal patch size depends on dataset size: larger patches benefit smaller datasets, while smaller patches are better for larger ones. Increasing the training set size generally enhances performance, especially with larger patches, by improving pattern learning and reducing overfitting. A balanced validation sample size is crucial for effective hyperparameter tuning, as very small or very large sets can lead to suboptimal results or excessive computation. Finally, very small test sets may not reliably estimate WaveMamba's generalization capability.

\begin{table}[!hbt]
    \centering
    \caption{Performance of WaveMamba with Different Patch Sizes and Various Training Percentages (\%)}
    \resizebox{\columnwidth}{!}{\begin{tabular}{c|cccc||c|cccc} \hline 
        \textbf{Train \%} & \textbf{AA} & \textbf{OA} & \textbf{$\kappa$} & \textbf{Tr Time} & \textbf{Patch Size} & \textbf{AA} & \textbf{OA} & \textbf{$\kappa$} & \textbf{Tr Time (s)} \\ \hline
        
        \multicolumn{10}{c}{\textbf{Pavia University}} \\ \hline 
        5\% & 93.11 & 95.79 & 94.43 & 445.35 &  $2 \times 2$ & 93.85 & 95.64 & 94.21 & 1407.88 \\
        10\% & 95.63 & 97.39 & 96.54 & 805.27 & $4 \times 4$ & 97.17 & 98.15 & 97.55 & 1586.50 \\
        15\% & 96.58 & 97.85 & 97.15 & 1064.72 & $6 \times 6$ & 96.42 & 97.51 & 96.70 & 1764.15 \\
        20\% & 97.13 & 98.19 & 97.61 & 1704.64 &  $8 \times 8$ & \textbf{97.87} & \textbf{98.67} & \textbf{98.23} & 1917.34 \\
        25\% & \textbf{97.87} & \textbf{98.67} & \textbf{98.23} & 1917.34 & $10 \times 10$ & 95.75 & 97.38 & 96.53 & 2149.22 \\ \hline
        
        \multicolumn{10}{c}{\textbf{University of Houston}} \\ \hline 
        5\% & 88.43 & 91.26 & 90.54 & 98.83 &  $2 \times 2$ & 97.01 & 97.57 & 97.38 & 1003.38 \\
        10\% & 94.98 & 96.18 & 95.87 & 199.04 &  $4 \times 4$ & \textbf{98.15} & \textbf{98.48} & \textbf{98.35} & 1043.73 \\
        15\% & 96.22 & 96.92 & 96.66 & 293.63 &  $6 \times 6$ & 96.37 & 97.39 & 97.17 & 1103.58 \\
        20\% & 96.96 & 97.71 & 97.52 & 444.91 &  $8 \times 8$ & 97.96 & 97.99 & 97.82 & 747.63 \\
        25\% & \textbf{98.15} & \textbf{98.48} & \textbf{98.35} & 1043.73 & $10 \times 10$ & 97.41 & 97.60 & 97.41 & 864.75 \\ \hline
    \end{tabular}}
    \label{Tab2}
\end{table}

\subsection{Impact of Wavelet Transformation}

We examined the impact of the wavelet transformation by comparing the model’s performance with and without wavelets, highlighting its contribution to feature enhancement as shown in Table  \ref{tab:table21} which provides a more detailed analysis of how the wavelets module’s absence affects overall performance, specifically focusing on performance degradation when key components like the wavelet transformation are removed. This will help emphasize the importance of each component in the model's architecture.

\begin{table}[!hbt]
    \centering
    \caption{Performance with and without Wavelet.}
    \resizebox{\columnwidth}{!}{\begin{tabular}{c|c|c|c|c|c} \hline
        \multirow{2}{*}{Class} & \multicolumn{2}{c|}{University of Houston} & \multirow{2}{*}{Class} & \multicolumn{2}{c}{Pavia University} \\ \cline{2-3} \cline{5-6}
        & Without & With & & Without & With \\ \hline 
        
        Healthy grass & 94.4089 & 99.0415 & Asphalt & 96.6224 & 98.5223 \\ 
        Stressed grass & 99.8405 & 99.5215 & Meadows & 99.3457 & 99.8712 \\ 
        Synthetic grass & 99.1379 & 100 & Gravel & 88.9418 & 92.2783 \\ 
        Trees & 96.6237 & 98.7138 & Trees & 97.4543 & 98.5639 \\ 
        Soil & 98.5507 & 99.8389 & Painted & 100 & 99.1084 \\ 
        Water & 93.8271 & 99.3827 & Soil & 97.4940 & 99.4431 \\ 
        Residential & 92.4290 & 93.6908 & Bitumen & 97.8947 & 98.1954 \\ 
        Commercial & 92.4437 & 95.8199 & Bricks & 92.3411 & 95.2743 \\ 
        Road & 93.6102 & 96.0063 & Shadows & 95.7805 & 98.1012 \\ \cline{4-6}
        Highway & 96.5798 & 98.8599 & \textbf{OA} & 96.2083 & \textbf{98.6347} \\ 
        Railway & 97.0873 & 98.0582 & \textbf{AA} & 97.3536 & \textbf{97.7065} \\ 
        Parking Lot 1 & 94.1653 & 98.8654 & \textbf{kappa} & 96.4942 & \textbf{98.1901} \\  \cline{4-6}
        Parking Lot 2 & 64.9572 & 88.0341 \\ 
        Tennis Court & 98.5981 & 99.5327 \\ 
        Running Track & 99.0909 & 100 \\ \cline{1-3}
        \textbf{OA} & 94.9833 & \textbf{97.8043} \\ 
        \textbf{AA} & 94.5737 & \textbf{97.6910} \\
        \textbf{kappa} & 94.0900 & \textbf{97.6260} \\  \cline{1-3}
    \end{tabular}}
    \label{tab:table21}
\end{table}
\begin{table*}[!hbt]
    \centering
    \caption{Per Class classification accuracy (Overall (OA) and Average Accuracy (AA) \%) and Kappa ($\kappa$) measure for \textbf{University of Houston}. (WM = WaveMamba).}
    \resizebox{\textwidth}{!}{\begin{tabular}{c|cccccccccccccccccccccc} \hline 
        \textbf{Class} & \cite{hong2022spectralformer} & \cite{9766028} & \cite{9926105} & \cite{10604879} & \cite{9785505} & \cite{10463068} & \cite{rs16060970} & \cite{9772356} & \cite{10443948} & \cite{9547387} & \cite{rs12030582} & \cite{9684381} & \cite{9864609} & \cite{9903640} & \cite{rs16132449} & \textbf{WM} \\ \hline
        
        1 & 93.31 & 97.26 & 93.39 & 97.24 & 83.00 & 88.13 & 81.23 & 80.51 & 79.47 & 93.44 & 93.48 & 95.36 & 92.67 & 90.90 & 92.88 & 99.04 \\
        2 & 97.81 & 97.29 & 99.54 & 97.69 & 83.74 & 90.04 & 87.59 & 67.45 & 79.08 & 95.37 & 95.10 & 98.58 & 97.57 & 95.11 & 95.99 & 99.52 \\
        3 & 100.0 & 98.74 & 100.0 & 97.76 & 89.70 & 100.0 & 88.19 & 99.12 & 100.0 & 98.67 & 100.0 & 99.20 & 98.17 & 97.10 & 100.0 & 100.0 \\
        4 & 100.0 & 95.78 & 98.09 & 97.85 & 92.57 & 89.49 & 77.05 & 97.39 & 98.51 & 97.69 & 97.13 & 97.82 & 96.58 & 92.27 & 99.17 & 98.71 \\
        5 & 98.16 & 98.41 & 97.70 & 99.91 & 99.81 & 98.20 & 90.09 & 96.64 & 100.0 & 98.61 & 97.66 & 99.99 & 99.78 & 98.58 & 98.29 & 99.83 \\
        6 & 100.0 & 91.36 & 100.0 & 92.46 & 100.0 & 99.30 & 93.94 & 84.66 & 87.53 & 95.08 & 97.37 & 98.42 &  97.47 & 96.42 & 96.39 & 99.38 \\
        7 & 87.83 & 94.60 & 90.96 & 90.70  & 87.59 & 98.97 & 87.53 & 78.78 & 77.00 & 91.46 & 91.93 & 84.32 & 86.43 & 89.18 & 91.60 & 93.69 \\
        8 & 85.91 & 91.82 & 89.18 & 96.78 & 73.22 & 98.01 & 93.02 & 45.49 & 49.77 & 78.02 & 94.88 & 80.80 & 83.25 & 83.90 & 83.33 & 95.81 \\
        9 & 75.33 & 92.39 & 90.62 & 94.85 & 81.21 & 95.94 & 86.12 & 76.05 & 72.19 & 92.26 & 88.57 & 78.82 & 82.61 & 86.58 & 92.05 & 96.01 \\
        10 & 82.52 & 90.61 & 93.22 & 99.27 & 68.15 & 98.65 & 78.68 & 36.93 & 94.51 & 97.61 & 89.76 & 95.10 & 96.50 & 99.09 & 98.05 & 98.85 \\
        11 & 79.19 & 89.09 & 87.91 & 96.13 & 89.85 & 87.38 & 81.65 & 64.05 & 85.59 & 93.82 & 95.44 & 95.82 & 93.46 & 92.34 & 92.48 & 98.05 \\
        12 & 72.76 & 94.18 & 83.15 & 99.18 & 89.15 & 94.33 & 89.71 & 70.01 & 59.63 & 92.45 & 93.15 & 89.65 & 89.55 & 91.27 & 91.24 & 98.86 \\
        13 & 79.49 & 82.51 & 84.11 & 65.16 & 92.28 & 98.95 & 85.49 & 98.81 & 80.06 & 95.86 & 82.75 & 97.75 & 92.98 & 91.22 & 92.87 & 88.03 \\
        14 & 93.90 & 91.55 & 97.21 & 98.96 & 100.0 & 98.79 & 93.02 & 99.81 & 100.0 & 99.95 & 98.13 & 99.85 & 99.88 & 100.0 & 100.0 & 99.53 \\
        15 & 97.50 & 96.72 & 100.0 & 100.0 & 100.0 & 62.16 & 88.33 & 96.41 & 99.57 & 99.45 & 99.05 & 100.0 & 99.97 & 98.63 & 100.0 & 100.0 \\ \hline  
        \textbf{OA} & 88.45 & 93.06 & 93.09 & 96.08 & 86.30 & 93.21 & 86.50 & 75.65 & 82.14 & 93.92 & 93.67 & 92.88 & 92.85 & 92.73 & 94.30 & \textbf{97.80} \\ \hline 
        \textbf{AA} & 87.81 & 86.61 & 92.06 & 94.93 & 88.68 & 93.22 & 85.25 & 79.34 & 84.19 & 94.65 & 94.03 & 94.10 & 93.79 & 93.51 & 94.96 & \textbf{97.69} \\ \hline 
        \textbf{$\kappa$} & 87.50 & 92.50 & 92.53 & 95.76  & 85.14 & 92.63 & 85.85 & 73.73 & 80.70 & 93.43 & 93.16 & 92.30 & 92.27 & 92.14 & 93.84 & \textbf{97.62} \\ \hline
    \end{tabular}}
    \label{Tab6}
\end{table*}
\begin{table*}[!hbt]
    \centering
    \caption{Per Class classification accuracy (Overall (OA) and Average Accuracy (AA) \%) and Kappa ($\kappa$) measure for \textbf{Pavia University}. (WM = WaveMamba).}
    \resizebox{\textwidth}{!}{\begin{tabular}{c|cccccccccccccccccccccc} \hline 
        \textbf{Class} & \cite{hong2022spectralformer} & \cite{9766028} & \cite{9926105} & \cite{10604879} & \cite{9785505} & \cite{10463068} & \cite{rs16060970} & \cite{9772356} & \cite{10443948} & \cite{9547387} & \cite{rs12030582} & \cite{9684381} & \cite{9864609} & \cite{9903640} & \cite{rs16132449} & \textbf{WM} \\ \hline
        
        1 & 92.67 & 95.21 & 95.25 & 96.34  & 90.44 & 94.75 & 98.76 & 78.09 & 76.94 & 97.06 & 98.74 & 86.16 & 87.16 & 90.59 & 95.70 & 98.52 \\
        2 & 92.79 & 92.54 & 96.64 & 99.51  & 96.63 & 95.54 & 95.46 & 84.67 & 93.70 & 92.29 & 99.51 &9 4.50 & 95.05 & 94.62 & 94.05 & 99.87 \\
        3 & 90.60 & 91.18 & 81.74 & 82.74 & 93.50 & 99.83 & 87.75 & 53.47 & 81.10 & 99.97 & 90.81 & 93.94 & 93.95 & 94.39 & 99.61 & 92.27 \\
        4 & 98.15 & 97.21 & 95.50 & 96.33  & 90.94 & 93.99 & 91.37 & 75.09 & 78.42& 97.12 & 92.74 & 88.97 & 92.76 & 84.61 & 98.92 & 98.56 \\
        5 & 98.28 & 100.0 & 99.59 & 100.0  & 98.71 & 95.82 & 92.13 & 99.83 & 100.0 & 100.0 & 99.53 & 98.95 & 98.85 & 99.18 & 100.0 & 99.11 \\
        6 & 93.29 & 99.93 & 93.43 & 92.31  & 98.82 & 97.32 & 97.10 & 52.40 & 64.24 & 99.40 & 90.96 & 96.07 & 99.19 & 99.53 & 99.19 & 99.44 \\
        7 & 83.01 & 95.75 & 89.24 & 92.06  & 98.13 & 96.58 & 96.47 & 78.74 & 96.27 & 100.0 & 93.80 & 99.58 & 99.08 & 99.34 & 99.93 & 98.19 \\
        8 & 84.50 & 97.59 & 88.81 & 90.67  & 93.95 & 95.37 & 86.25 & 84.35 & 58.58 & 98.99 & 89.83 & 86.63 & 91.74 & 97.10 & 98.50 & 95.27 \\
        9 & 99.77 & 99.47 & 99.53 & 93.30  & 95.24 & 92.18 & 89.27 & 99.60 & 98.16 & 99.47 & 96.82 & 95.54 & 95.49 & 93.95 & 99.96 & 98.10 \\ \hline 
        
        \textbf{OA} & 92.30 & 91.35 & 94.48 & 96.05  & 95.25 & 95.05 & 94.64 & 78.23 & 83.28 & 95.79 & 95.87 & 92.61 & 94.06 & 94.33 & 96.40 & \textbf{98.63} \\ \hline 
        \textbf{AA} & 88.86 & 85.07 & 93.15 & 93.71  & 95.15 & 93.01 & 92.78 & 78.47 & 83.06 & 98.25 & 94.75 & 93.37 & 94.81 & 94.81 & \textbf{98.43} & 97.70 \\ \hline 
        \textbf{$\kappa$} & 89.66 & 88.94 & 92.67 & 94.74  & 93.55 & 94.70 & 93.54 & 71.55 & 77.74 & 94.54 & 94.58 & 90.29 & 92.22 & 92.60 & 95.31 & \textbf{98.19} \\ \hline
    \end{tabular}}
    \label{Tab4}
\end{table*}

\subsection{Experimental Results with CNN-based Models}

\begin{figure}[!hbt]
    \centering
	\includegraphics[width=0.49\textwidth]{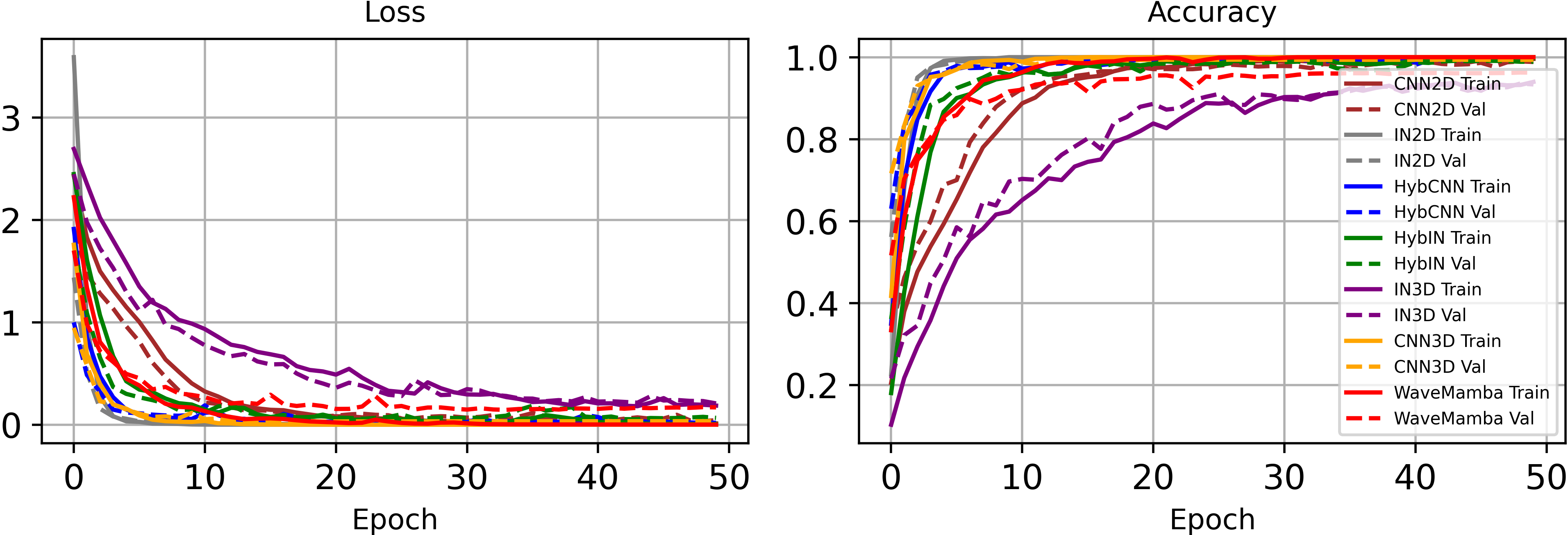}
    \caption{Accuracy and loss trends of CNN-based methods.}
\label{FigAcc}
\end{figure}

\begin{figure}[!hbt]
    \centering
	\begin{subfigure}{0.15\textwidth}
		\includegraphics[width=0.99\textwidth]{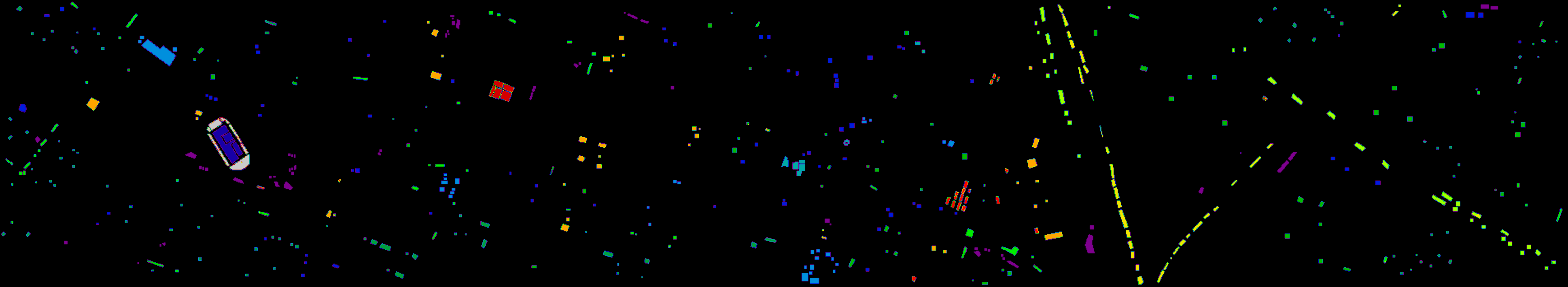}
		\caption{2DCNN} 
		\label{Fig3A}
	\end{subfigure}
	\begin{subfigure}{0.15\textwidth}
		\includegraphics[width=0.99\textwidth]{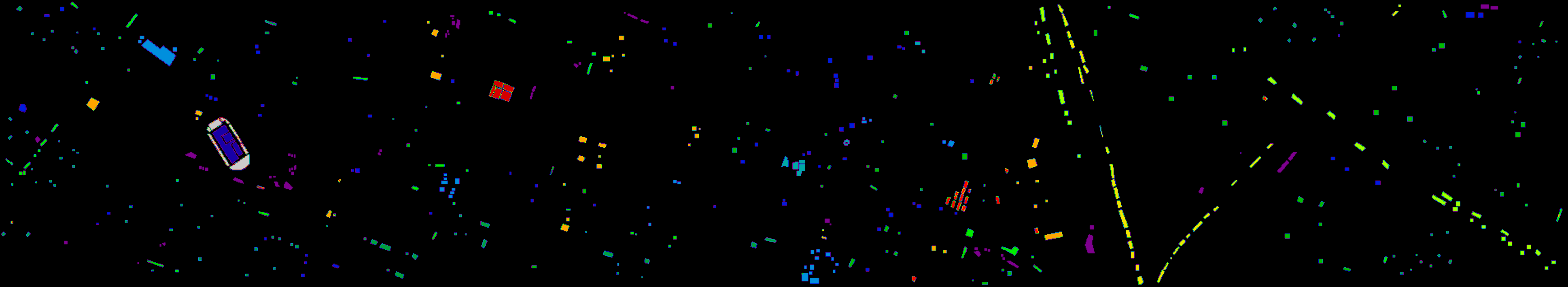}
		\caption{3DCNN}
		\label{Fig3B}
	\end{subfigure} 
	\begin{subfigure}{0.15\textwidth}
		\includegraphics[width=0.99\textwidth]{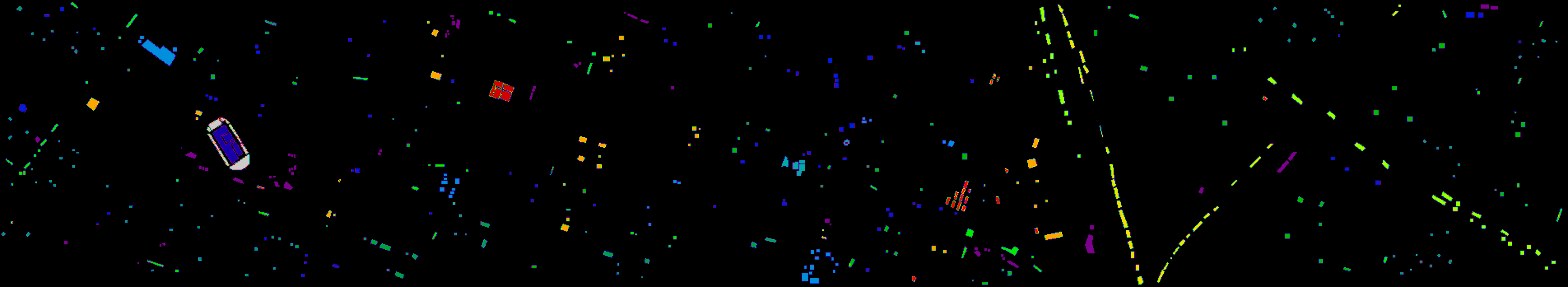}
		\caption{HybridCNN}
		\label{Fig3C}
	\end{subfigure} 
	\begin{subfigure}{0.15\textwidth}
		\includegraphics[width=0.99\textwidth]{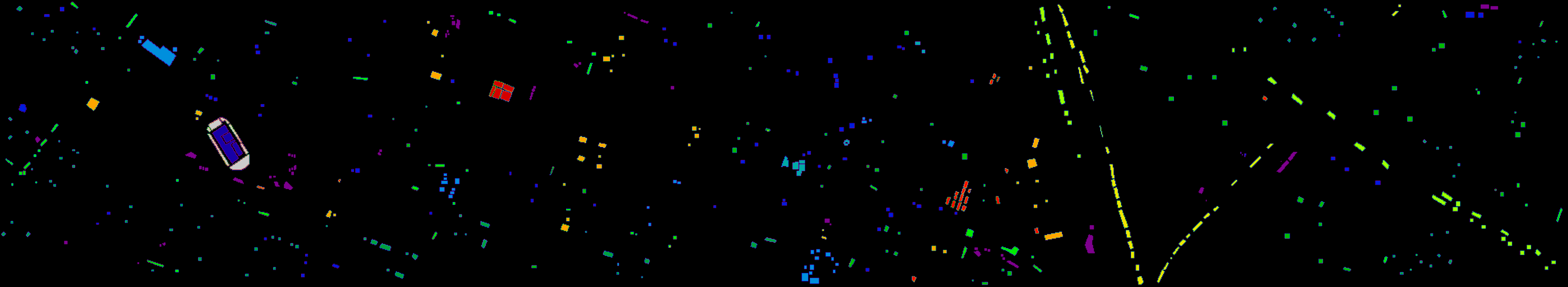}
		\caption{2DIN}
		\label{Fig3D}
	\end{subfigure} 
	\begin{subfigure}{0.15\textwidth}
		\includegraphics[width=0.99\textwidth]{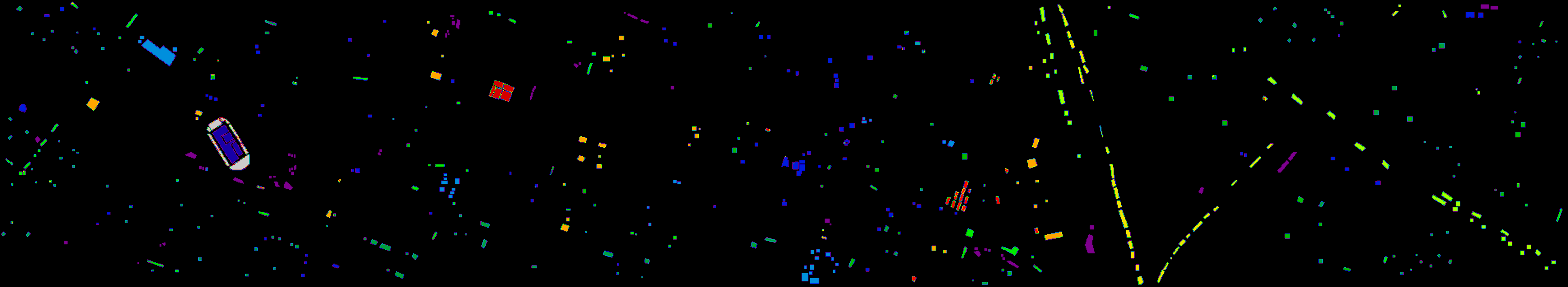}
		\caption{3DIN}
		\label{Fig3E}
	\end{subfigure} 
	\begin{subfigure}{0.15\textwidth}
		\includegraphics[width=0.99\textwidth]{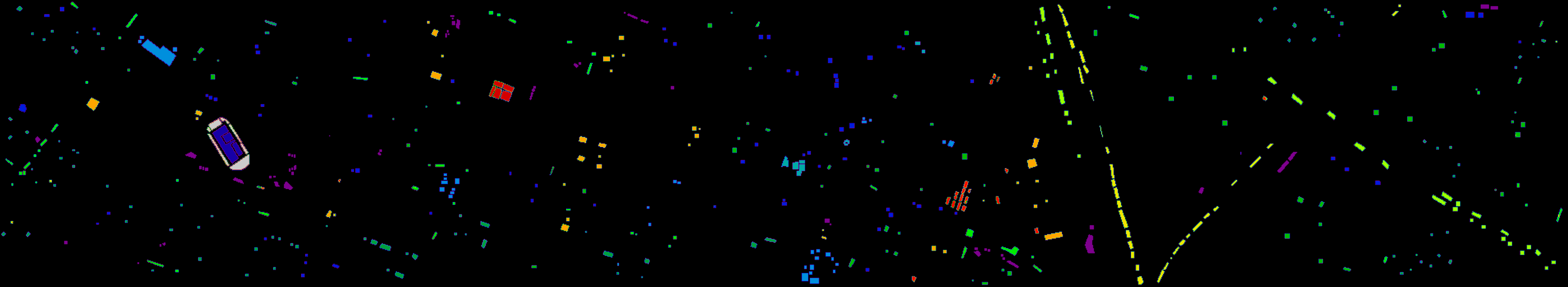}
		\caption{HybridIN}
		\label{Fig3F}
	\end{subfigure} 
    \begin{subfigure}{0.15\textwidth}
		\includegraphics[width=0.99\textwidth]{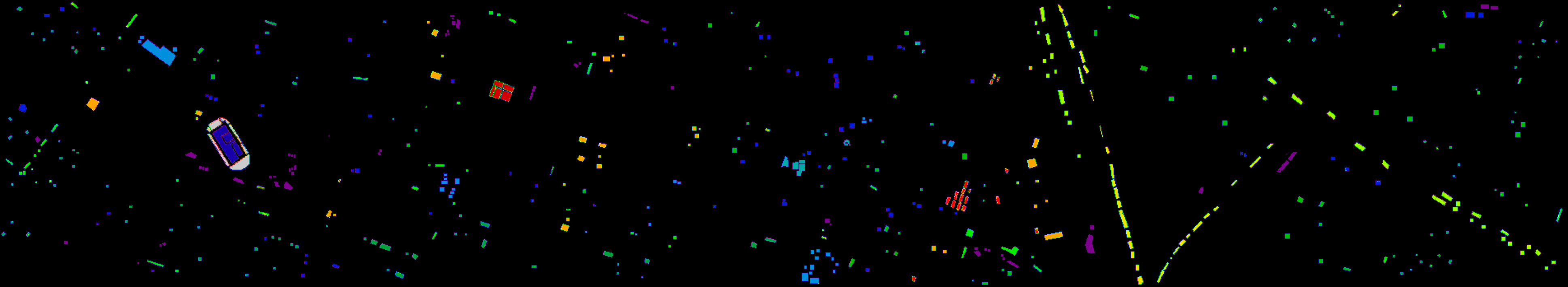}
		\caption{WaveMamba}
		\label{Fig3G}
	\end{subfigure}
\caption{The proposed WaveMamba achieves OA=97.30\% showing competitive performance. All these results are compiled using $10 \times 10$ patch size with 25\% training samples for all competing methods.}
\label{Fig3}
\end{figure}

\begin{figure}[!hbt]
    \centering
	\begin{subfigure}{0.06\textwidth}
		\includegraphics[width=0.99\textwidth]{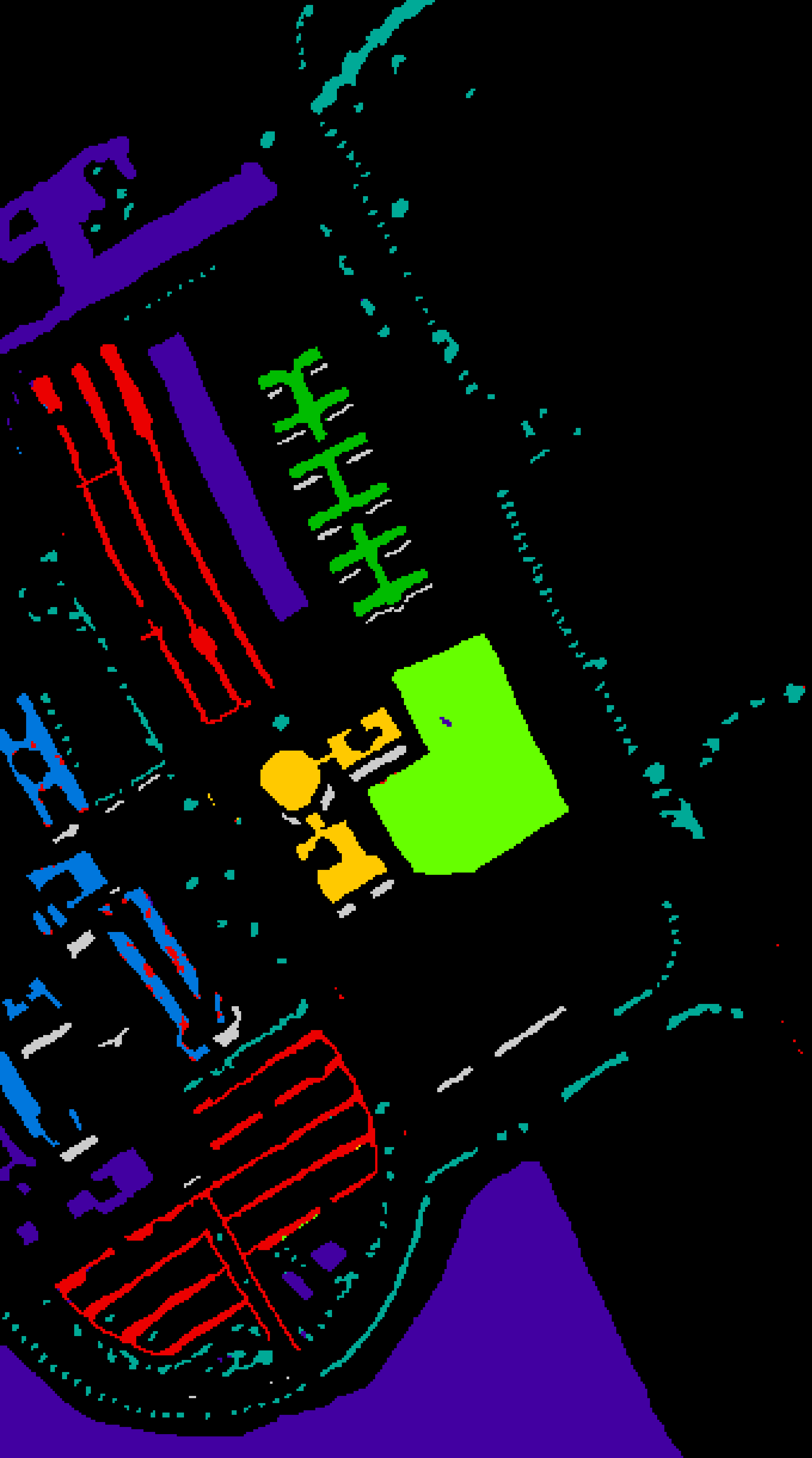}
		\caption{2DCNN} 
		\label{Fig7A}
	\end{subfigure}
	\begin{subfigure}{0.06\textwidth}
		\includegraphics[width=0.99\textwidth]{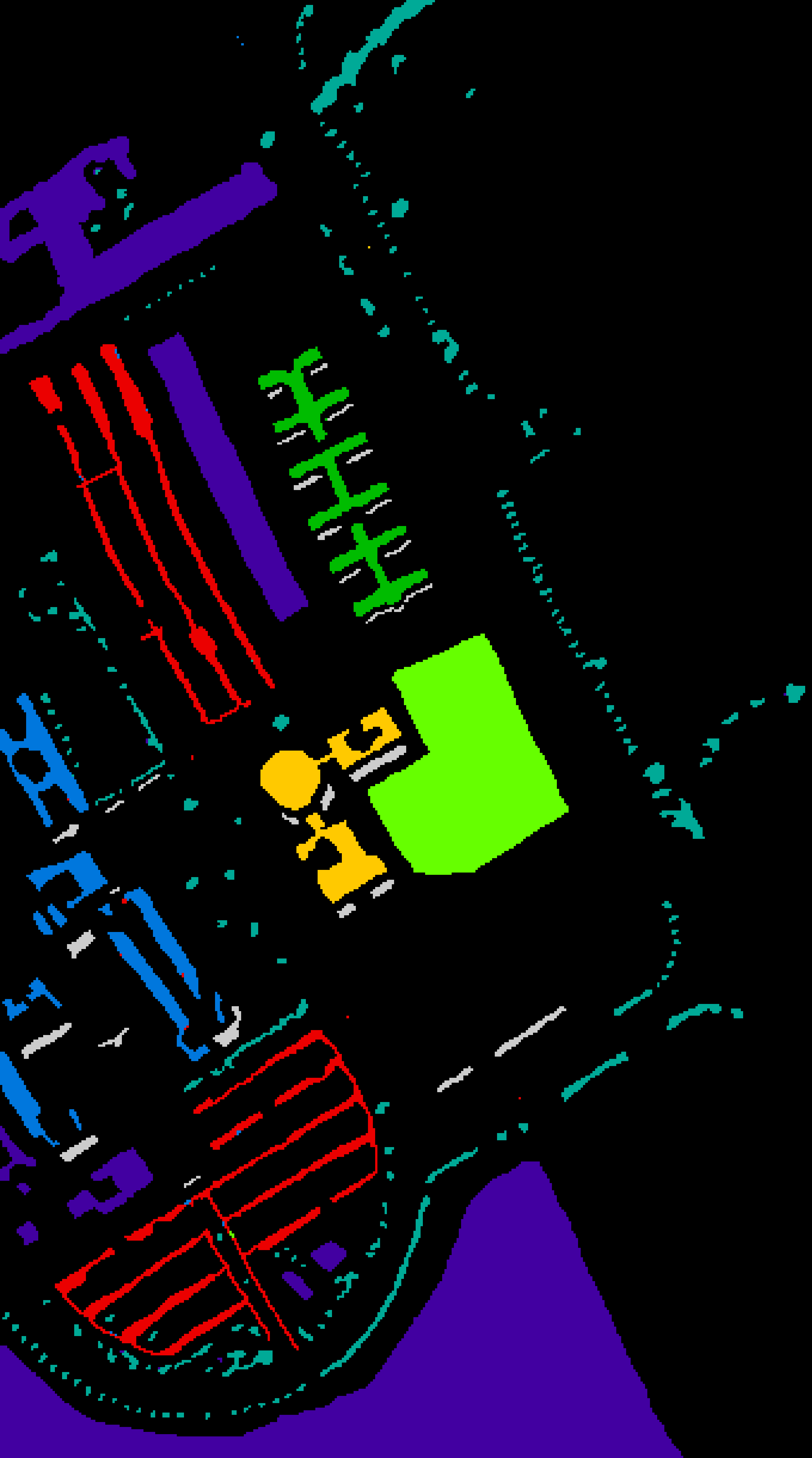}
		\caption{3DCNN}
		\label{Fig7B}
	\end{subfigure}
	\begin{subfigure}{0.06\textwidth}
		\includegraphics[width=0.99\textwidth]{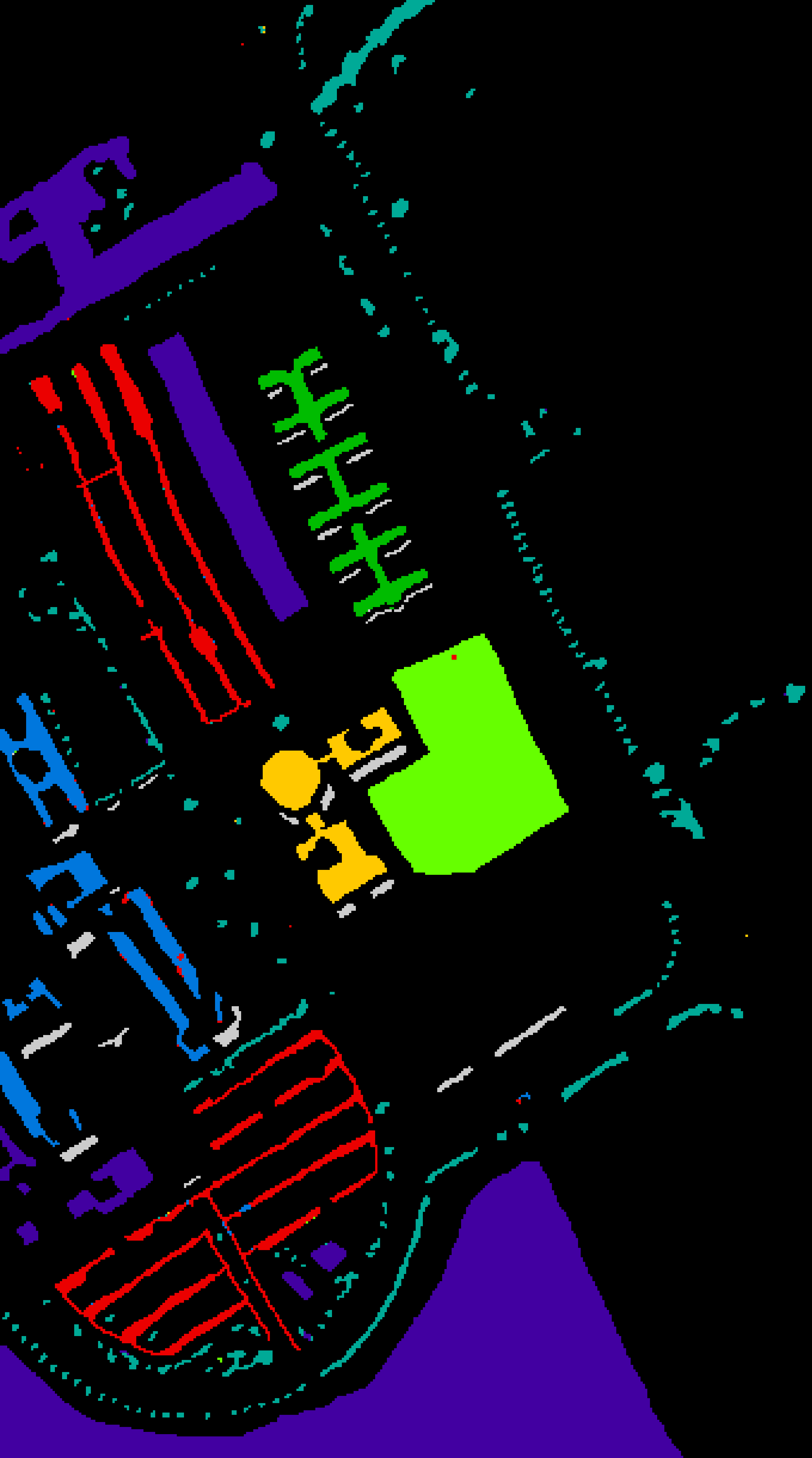}
		\caption{HybCNN}
		\label{Fig7C}
	\end{subfigure}
	\begin{subfigure}{0.06\textwidth}
		\includegraphics[width=0.99\textwidth]{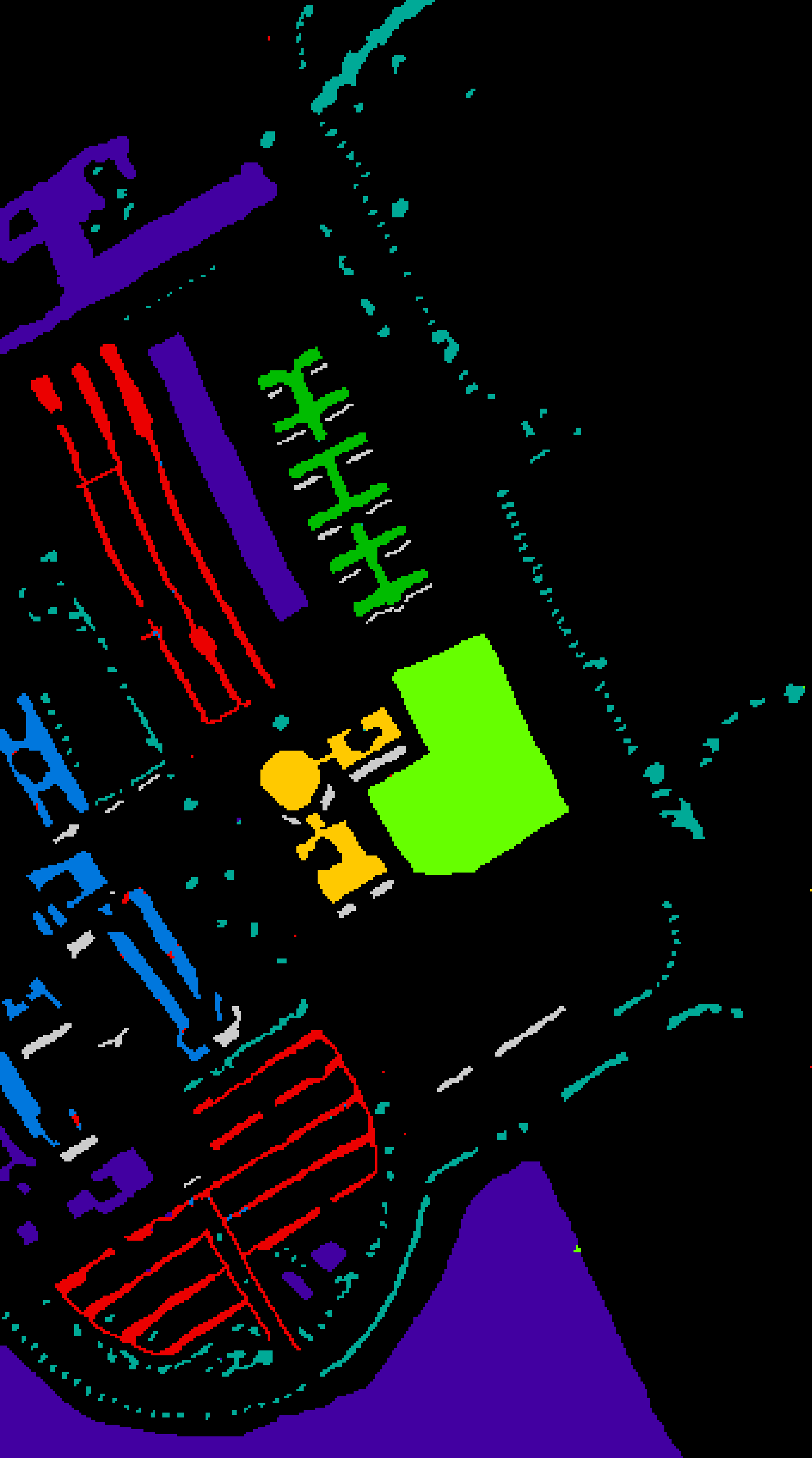}
		\caption{2DIN}
		\label{Fig7D}
	\end{subfigure}
	\begin{subfigure}{0.06\textwidth}
		\includegraphics[width=0.99\textwidth]{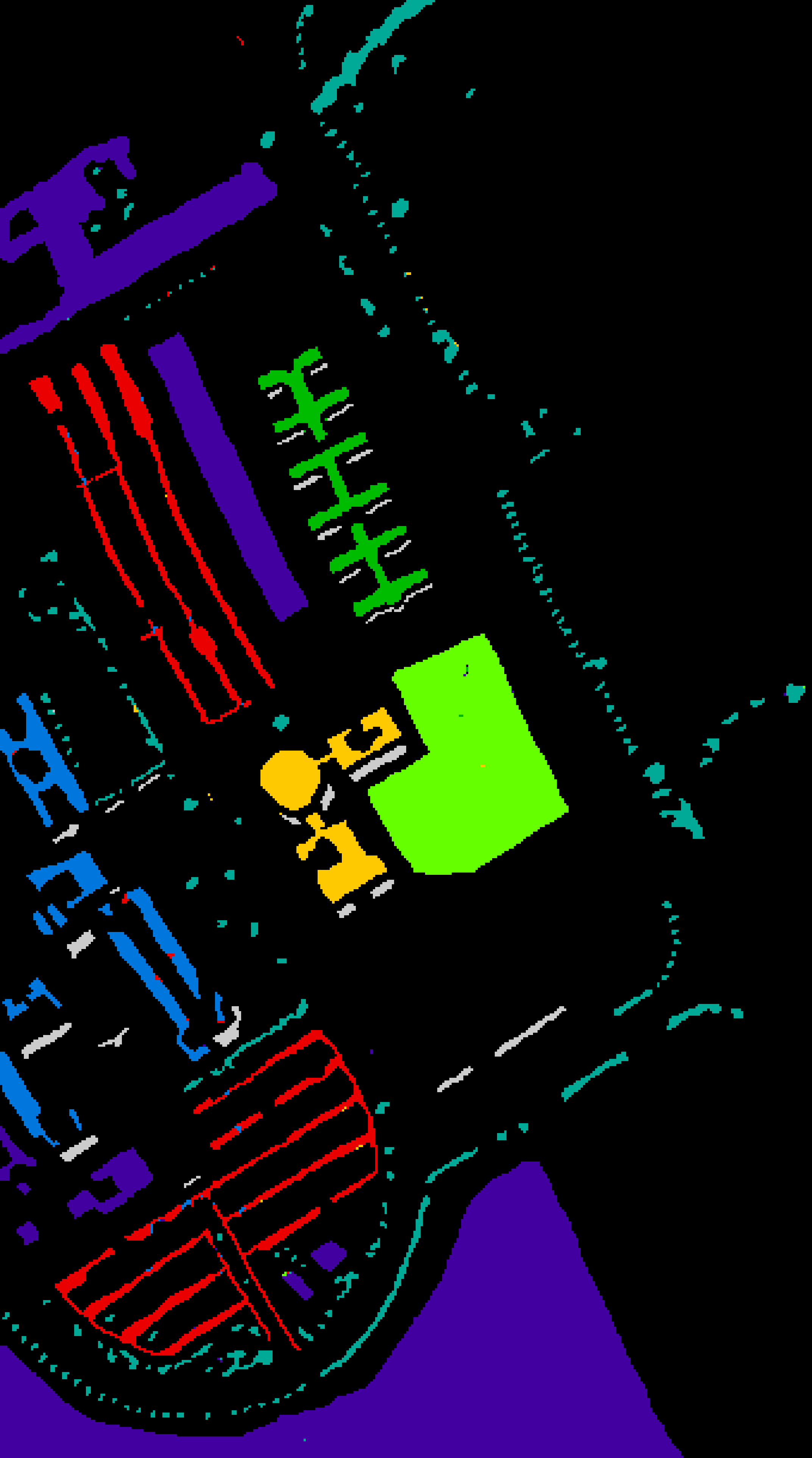}
		\caption{3DIN}
		\label{Fig7E}
	\end{subfigure} 
	\begin{subfigure}{0.06\textwidth}
		\includegraphics[width=0.99\textwidth]{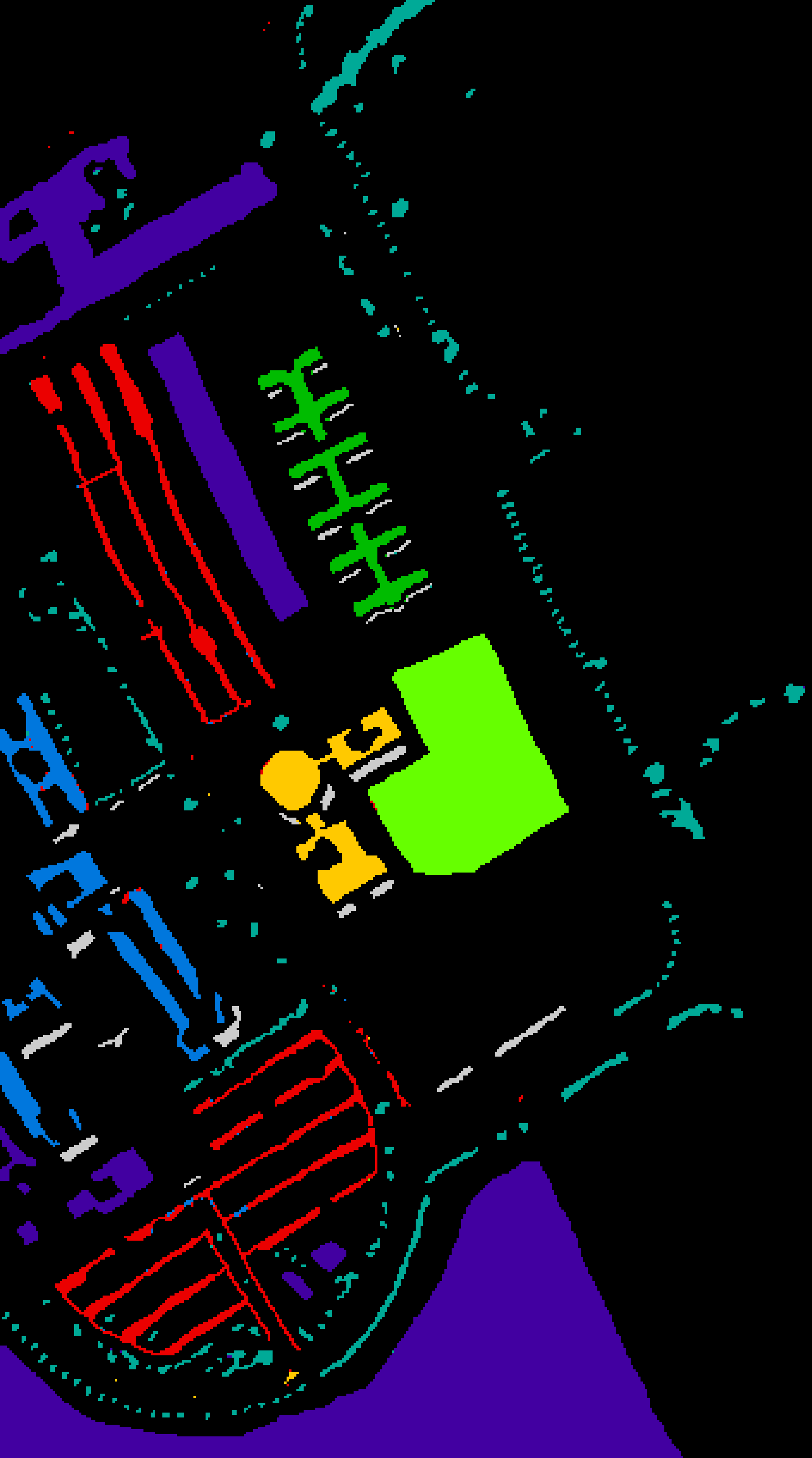}
		\caption{HybIN}
		\label{Fig7F}
	\end{subfigure} 
    \begin{subfigure}{0.06\textwidth}
		\includegraphics[width=0.99\textwidth]{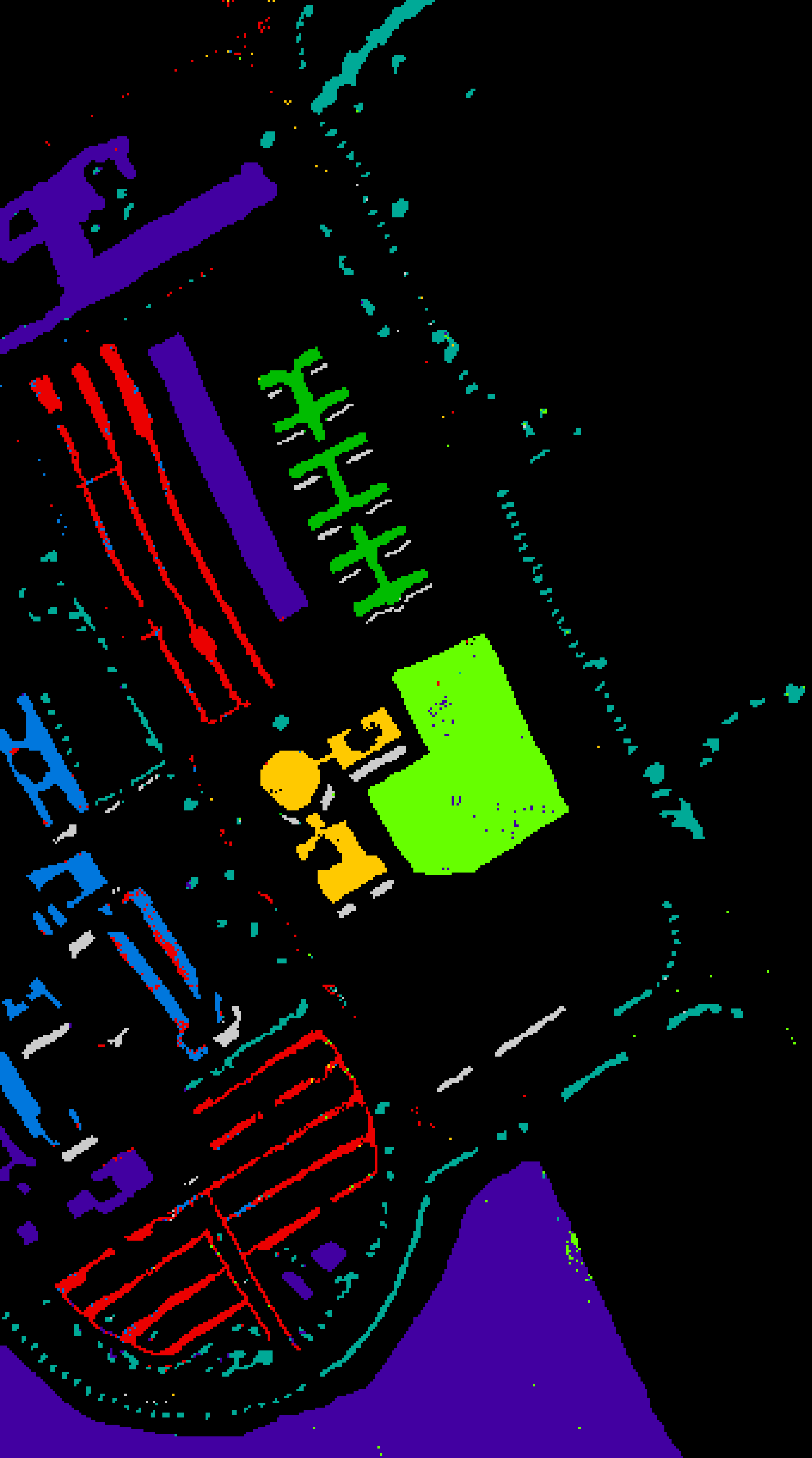}
		\caption{WMamba}
		\label{Fig7G}
	\end{subfigure} 
\caption{The proposed WaveMamba achieves OA=96.58\% showing competitive performance. All these results are compiled using $10 \times 10$ patch size with 25\% training samples for all competing methods.}
\label{Fig7}
\end{figure}

A consistent experimental methodology was employed for fair model comparison using unique sample distributions, spatial locations, and a controlled $10 \times 10$ pixel patch size. Figure \ref{FigAcc} illustrates the loss and accuracy trends on the University of Houston dataset, while ground truth maps for WaveMamba and other models are shown in Figures \ref{Fig3} and \ref{Fig7}. The evaluated models include 2D CNN, 3D CNN, Hybrid Inception Net, 3D Inception Net, 2D Inception Net, and Hybrid CNN. The results demonstrate the benefits of decoupling spatial and spectral information with Wavelets in WaveMamba, which achieved impressive scores of 97\%, comparable to CNN models. The marginal accuracy difference between WaveMamba and other CNN-based methods is likely due to WaveMamba's computational efficiency, reduced overfitting, and improved modeling of spatial and spectral dependencies, with similar trends observed for the Pavia University dataset.


\subsection{Comparative Results with Transformer and Mamba}

To illustrate the effectiveness of WaveMamba, it was compared against a range of HSI classification methods, including Spectralformer (SF) \cite{hong2022spectralformer}, HiT \cite{9766028}, CSiT \cite{9926105}, SSFTT \cite{9684381}, S3L \cite{rs16060970}, RIAN \cite{9785505}, CAT \cite{10463068}, SPRLT \cite{9772356}, CMT \cite{10443948}, and additional models such as DBDA \cite{rs12030582}, MSSG \cite{9547387}, LSFAT \cite{9864609}, CT-Mixer \cite{9903640}, SSFormer \cite{10604879}, and SS-Mamba \cite{rs16132449}. This diverse set includes advanced CNN models incorporating dual-attention mechanisms, Transformer-based approaches, and hybrid models, thereby providing a comprehensive benchmark for assessing WaveMamba's performance.

The results, as presented in Tables \ref{Tab4} and \ref{Tab6}, demonstrate that WaveMamba surpasses the competing methods in terms of Overall Accuracy (OA), Average Accuracy (AA), and Kappa ($\kappa$) measure on both datasets. Specifically, WaveMamba achieves OA values of 97.80\% and 98.63\% for the respective datasets, outperforming the second-best method, SS-Mamba, by margins of 0.18\% and 2.09\%. Furthermore, WaveMamba exhibits superior performance in AA and $\kappa$ measure, underscoring its robustness and efficacy. Notable improvements are observed in classes 3, 6, 8, and 9 on the Pavia University dataset, and in classes 8, 10, 12, and 15 on the University of Houston dataset. These results highlight WaveMamba's capability to accurately classify a variety of land cover types, affirming its potential as a promising approach.

\section{Conclusions}
\label{Con}

This paper introduced "WaveMamba," which integrates wavelet transformation with the Spatial-Spectral Mamba architecture for HSI classification. By leveraging wavelet-based extraction of multi-scale spatial-spectral features, WaveMamba captures both local texture patterns and global contextual relationships in an end-to-end trainable model. A key innovation is the incorporation of a State-Space Model, enhancing the modeling of temporal dependencies alongside spatial-spectral information. Extensive experiments show that WaveMamba significantly improves classification accuracy over well-known advanced DL methods, especially in datasets with limited training data, by effectively capturing nuanced spectral and structural details. The method not only achieves superior performance but also demonstrates robustness and generalizability, making it a promising solution for real-world HSI classification challenges. Future work could focus on exploring self-supervised pre-training and further network optimizations to enhance WaveMamba's performance in data-scarce environments.

\bibliographystyle{IEEEtran}
\bibliography{IEEEabrv,Sam}
\end{document}